\documentclass[sigconf, nonacm]{acmart}
\usepackage{booktabs}
\usepackage{multirow}
\usepackage{amsmath}

\usepackage{amssymb}
\usepackage{mathtools}
\usepackage[T1]{fontenc}
\renewcommand\footnotetextcopyrightpermission[1]{}  % no copyright footnote
\acmConference[KDD 2026 Workshop]{Enterprise AI Agents: From 
Prototypes to Production}{August 9--13, 2026}{Jeju, Korea}

\begin{document}

% \title{Co-Evolutionary Adversarial Training for Self-Improving Customer Support Agents Without Human Feedback}
\title{RL-ADA: A World-Feedback Framework for Adversarially Robust 
Enterprise Dialogue Agents}

%% --- authors (RLEval is single-blind; replace placeholders) ---
\author{Ram Narayanan}
\affiliation{
\country{}
% \country{USA}
  \institution{Centific}}
\email{ram.sampath@centific.com}

\author{Harshit Rajgarhia}
\affiliation{
\country{}
% \country{USA}
  \institution{Centific}}
\email{harshit.rajgarhia@centific.com}

\author{Abhishek Mukherji}
\affiliation{%
\country{}
% \country{USA}
  \institution{Centific}}
\email{abhishek.mukherji@centific.com}

\renewcommand{\shortauthors}{Narayanan, Rajgarhia and Mukherji}

\begin{abstract}
Deploying task-oriented dialogue agents in enterprise customer 
support faces a persistent annotation bottleneck: robust training 
requires labelled interaction data at scale, yet enterprise 
conversational logs are privacy-sensitive and expensive to annotate, 
while user behaviour evolves faster than labelling pipelines can 
keep pace. We present RL-ADA (Reinforcement Learning with 
Adversarial Dialogue Agents), a co-evolutionary training framework 
that eliminates this bottleneck by replacing human labels with 
\emph{world feedback}: consequence-based reward signals derived 
directly from measurable interaction outcomes. A Customer Support 
Agent (DA, 3B parameters) and an Adversarial Customer Agent 
(CA, 7B parameters) co-evolve in an adversarial arena guided by 
a fixed automated judge: the DA is rewarded for correctly handling 
multi-turn customer conversations to successful resolution, while 
the CA is rewarded for producing realistic, intent-concealing 
utterances that cause misroutes, creating asymmetric adversarial 
pressure through opposing but independently structured rewards. 
An isolation gym iteratively retrains the weaker agent on 
prior-failure transcripts, requiring no human annotation at any 
stage. In a banking customer support proof of concept, tool-routing 
errors are eliminated and the strict end-to-end PASS rate doubles 
over five co-evolutionary cycles, driven solely by automated arena 
reward with no labelled data. We additionally observe the emergence 
of \textbf{Contextual Camouflage}, an adversarial strategy in 
which the CA learns to embed intent within dense realistic customer 
detail purely from reward pressure, with direct implications for 
enterprise red-teaming and robustness evaluation.
\end{abstract}


\maketitle

%%%%%%%%%%%%%%%%%%%%%%%%%%%%%%%%%%%%%%%%%%%%%%%%%%%%%%%%%%%%%%%%%%%%%%

\section{Introduction}
\label{sec:intro}

Task-oriented dialogue agents are increasingly deployed in 
enterprise settings where they must handle unpredictable, 
adversarial, and continuously evolving user inputs. Training 
such agents to be robust requires large volumes of labelled 
interaction data, yet enterprise conversational logs are 
privacy-sensitive, domain-specific, and expensive to annotate. 
The result is a fundamental bottleneck: agents that perform 
well on evaluation benchmarks but degrade under the adaptive 
pressure of real users.

Existing approaches only partially address this bottleneck. 
RLHF~\cite{ouyang2022instructgpt} reduces agent training to a 
human-annotation problem that is costly and difficult to scale 
as user behaviour evolves. Self-play 
methods~\cite{silver2016mastering,berner2019dota} avoid 
annotation by learning from interaction outcomes, but assume 
symmetric agents with identical action spaces and train both 
simultaneously, inducing non-stationarity and risking 
catastrophic forgetting. Neither approach directly addresses the 
\emph{asymmetric, annotation-free} case native to customer 
support: a support agent that must correctly route tool calls 
versus an adversarial customer that elicits misroutes through 
natural language, each requiring different model capacities, 
action spaces, and reward structures.

We present \textbf{RL-ADA} (Reinforcement Learning with 
Adversarial Dialogue Agents), a framework that replaces human 
labels with \emph{world feedback}: consequence-based reward 
signals derived directly from measurable interaction outcomes. 
A 3B \textbf{Customer Support Agent} (DA) and a 7B 
\textbf{Adversarial Customer Agent} (CA), both implemented as 
language model policies, train against each other without any 
human annotation at any stage.

\paragraph{Contributions.}
\begin{itemize}
  \setlength{\itemsep}{1pt}
  \setlength{\parsep}{0pt}
  \setlength{\topsep}{2pt}
  \item A \textbf{co-evolutionary training loop} in which a 3B 
        DA language model and a 7B CA language model train 
        against each other using \emph{world feedback} only: 
        rule-based outcome signals combined with a fixed 
        automated judge whose reliability we evaluate against 
        GPT-4o-mini, with no human annotation at any stage 
        (\S\ref{sec:arch}, \S\ref{sec:judge}).
  \item An \textbf{isolation gym} that retrains the weaker 
      agent on a 70:30 failure/success transcript mix, 
      producing non-monotonic co-evolutionary dynamics 
      and providing the training-side mechanism for a 
      post-deployment data flywheel (\S\ref{sec:gym}, 
      \S\ref{sec:exp:arena}).
  \item A \textbf{macro-level stopping criterion} for asymmetric 
        adversarial training based on rolling win-rate stability, 
        framed as a practical surrogate for empirical 
        $\varepsilon$-Nash convergence (\S\ref{sec:stop}).
\end{itemize}

\paragraph{Instantiation.}
We instantiate RL-ADA on a banking customer support setting 
where the DA must route free-text customer utterances to one 
of six API tools across 78 distinct customer intents 
(Appendix~\ref{app:taxonomy}). Customers never name their 
intent directly: \emph{``I see something odd on my statement''} 
could require \texttt{dispute\_charge} or 
\texttt{get\_transactions}, and an adversarial customer can 
phrase requests to trigger the wrong tool. The DA is not an 
intent classifier and is trained only from call outcomes with 
no human labels.

%%%%%%%%%%%%%%%%%%%%%%%%%%%%%%%%%%%%%%%%%%%%%%%%%%%%%%%%%%%%%%%%%%%%%%
\section{Related Work}
\label{sec:related}

\paragraph{RLHF and AI feedback.}
Ouyang et al.~\cite{ouyang2022instructgpt} establish human 
preference feedback as the dominant alignment paradigm; Bai 
et al.~\cite{bai2022constitutional} partially displace it via 
RLAIF. RL-ADA replaces both with structured episode scores 
from interaction outcomes, eliminating annotation dependency 
at the cost of judge reliability uncertainty (\S\ref{sec:judge}).

\paragraph{Self-play and game-theoretic RL}
AlphaGo~\cite{silver2016mastering}, OpenAI 
Five~\cite{berner2019dota}, and 
AlphaStar~\cite{vinyals2019alphastar} demonstrate symmetric 
self-play at scale; PSRO~\cite{mcmahan2003double,lanctot2017unified} 
formalises iterative best-response over populations. RL-ADA's 
isolation gym is structurally the inner loop of Double Oracle 
with a single oracle per cycle and a curated failure/success 
curriculum rather than uniform population replay. 
Customer-agent dialogue is asymmetric by construction; 
simultaneous co-training under asymmetry induces cycling 
dynamics~\cite{mertikopoulos2018cycles,balduzzi2018mechanics} 
that RL-ADA's freeze-one-train-one structure avoids.

\paragraph{LLM red-teaming and concurrent work.}
Wallace et al.~\cite{wallace2019universal} and Perez et 
al.~\cite{perez2022redteaming} establish automated adversarial 
test generation against fixed targets; ARLAS~\cite{wang2025arlas} 
and SPAG~\cite{chen2024spag} apply adversarial self-play to 
LLM safety and reasoning. RL-ADA differs in that the 
Adversarial CA co-evolves with the DA rather than attacking a 
fixed model, targets tool-routing correctness in asymmetric 
dialogue, and uses asymmetric model sizes (3B vs.\ 7B) under 
an explicit convergence criterion.

\paragraph{Task-oriented dialogue evaluation.}
Yao et al.~\cite{yao2024taubench} and Barres et 
al.~\cite{barres2025tau2} establish tool-agent-user evaluation 
benchmarks with simulated cooperative users. RL-ADA differs 
in that the CA is adversarially trained to cause failures 
rather than simulate realistic task completion.
%%%%%%%%%%%%%%%%%%%%%%%%%%%%%%%%%%%%%%%%%%%%%%%%%%%%%%%%%%%%%%%%%%%%%%
\section{System Architecture}
\label{sec:arch}

\begin{figure}[t]
  \vspace{-6pt}
  \centering
  \includegraphics[width=\linewidth,height=0.30\textheight,keepaspectratio]{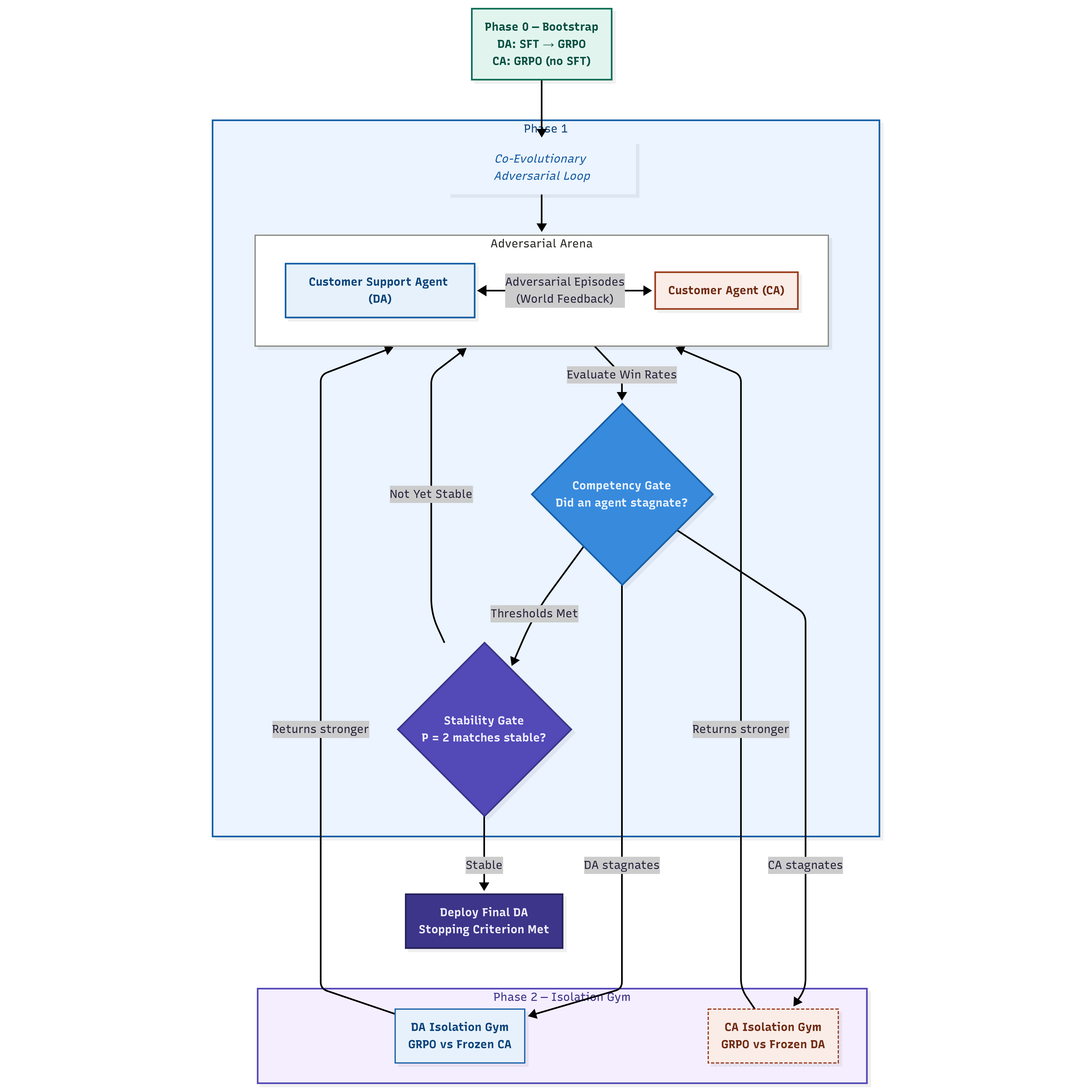}
  \caption{The RL-ADA three-phase loop. Bootstrap initialises
           both agents; the Arena measures relative win rates;
           the Isolation Gym retrains the weaker agent on
           failure transcripts.}
  \label{fig:loop}
  \vspace{-8pt}
\end{figure}

RL-ADA uses three model roles: a \textbf{Customer Support Agent} 
(referred to as DA throughout), an \textbf{Adversarial Customer Agent} 
(CA), and a \textbf{Judge}. The DA (Qwen2.5 3B) manages 
multi-turn clarification, routes tool calls, and ends calls; the CA 
(Qwen2.5 7B) is an RL-trained language model policy that generates 
customer-style utterances designed to elicit misroutes; the Judge 
(Qwen2.5 7B) provides terminal episode quality scores. Training 
proceeds across three phases (Figure~\ref{fig:loop}).

\subsection{Automated Judge}
\label{sec:judge}

Replacing human annotation requires a reliable automated signal 
for episode quality. Rather than using a proprietary model such 
as GPT-4o-mini as the judge, which would introduce an external 
dependency into the training loop, we use a
Qwen2.5-7B model (NeutralJudge) as a fixed episode scorer. 
The key question is whether a local 7B model is sufficiently 
reliable for this task: reliable enough in resolution detection, 
quality ranking, and hallucination detection to serve as the 
sole training signal without systematic bias that would corrupt 
the reward. We evaluate NeutralJudge against GPT-4o-mini as a 
reference point on $n{=}38$ conversations (8 synthetic with 
tier labels + 30 ABCD~\cite{chen2021abcd}; tier Spearman on 
the 8 synthetic only). NeutralJudge matches GPT-4o-mini in 
resolution detection (F1 0.807 vs.\ 0.821), ranks quality 
tiers in the same order ($\rho{=}0.833$), and is 
well-calibrated for relative episode ranking. Hallucination detection is particularly critical: a judge that misses fabricated facts would corrupt the reward signal, 
making F1 of 1.000 a hard prerequisite for safe RL training
(Table~\ref{tab:judge}).

\begin{table}[h]
  \vspace{-4pt}
  \centering
  \caption{NeutralJudge (7B) vs.\ GPT-4o-mini ($n{=}38$).}
  \label{tab:judge}
  \footnotesize
  \renewcommand{\arraystretch}{0.85}
  \begin{tabular}{lc}
    \toprule
    \textbf{Metric} & \textbf{Value} \\
    \midrule
    Resolution F1 (NeutralJudge / GPT-4o-mini) & 0.807 / 0.821 \\
    Quality-tier rank Spearman ($n{=}8$ synth.) & 0.833 \\
    Cross-judge overall Pearson                 & 0.779 \\
    Score calibration bias                      & $+$0.50 \\
    Hallucination detection F1                  & 1.000 \\
    \bottomrule
  \end{tabular}
  \vspace{-8pt}
\end{table}

\subsection{Phase 0 -- Bootstrap}

The DA is first trained via supervised fine-tuning on 
tool-routing demonstrations derived from 
Banking77~\cite{casanueva2020banking77}, then refined with 
GRPO~\cite{shao2024deepseekmath} using automated judge reward. 
The CA receives no SFT initialisation; it learns to produce 
misleading customer utterances from reward pressure against a 
fixed DA checkpoint.

The DA operates in a \emph{partially observable environment} 
implemented as an OpenEnv~\cite{openenv2026} 
\textsc{CustomerEnvironment}: at each turn it receives only 
the conversation history and must decide whether to call a 
tool, speak, or end the call, without ever observing the CA's 
hidden intent. The DA reasons explicitly before each action 
via a structured \texttt{thinking} field in its JSON output, 
performing implicit intent inference as part of its 
chain-of-thought before committing to a tool call. The 78 
customer \emph{intent} strings collapse to 6 tools via a 
many-to-one map (Appendix~\ref{app:taxonomy}): for example, 
\texttt{unrecognized\_charge} and 
\texttt{refund\_not\_showing\_up} both route to 
\texttt{dispute\_charge}. The reward is on the \emph{outcome}, 
not on any intermediate intent label, so the system requires 
no labelled annotations at training time.

\subsection{Reward Design}
\label{sec:reward}

The total DA reward per GRPO completion is
\begin{equation}
r_{\text{DA}} = r_{\text{format}} + r_{\text{tool}} + 
r_{\text{env}} + r_{\text{end}},
\label{eq:r_da}
\end{equation}
summarised in Table~\ref{tab:rewards}. $r_{\text{env}}$ is the 
return from a full \textsc{CustomerEnvironment} forward rollout 
combining turn-level rule-based signals with a terminal judge 
score clipped to $[-2,+2]$. Procedural failures (missed 
identity verification or missing domain tool) trigger mandatory 
negative deductions ($-1.2$ and $-1.8$ respectively), ensuring 
the signal is negative whenever the DA fails either required 
step regardless of other scores.

The CA reward is
\begin{equation}
r_{\text{CA}} = r_{\text{format}} + r_{\text{realism}} +
r_{\text{conceal}} + r_{\text{inv-DA}},
\label{eq:r_ca}
\end{equation}
where $r_{\text{format}}$ rewards natural-language output rather
than JSON, $r_{\text{realism}}$ is a signed shaping term favouring
utterances that sound like a real customer rather than a script,
$r_{\text{conceal}}$ penalises utterances that directly name the
intent, and $r_{\text{inv-DA}}$, the core signal, rewards misroutes
and penalises correct DA routing. Component weights 
in both reward functions were chosen pragmatically from early 
stability observations and are tunable per deployment domain; 
sensitivity analysis is left to future work (\S\ref{sec:limits}). 
The dense turn-level shaping terms bootstrap the action space; 
the terminal judge $J$ is the world-feedback signal that drives 
policy improvement once early-training mode collapse is 
prevented.

\begin{table}[t]
  \centering
 \caption{DA reward components. Values shown were chosen from 
           early stability observations and are tunable per 
           deployment domain (e.g.\ escalation cost, 
           hallucination tolerance).}
  % \caption{DA reward components.}
  \label{tab:rewards}
  \footnotesize
  \begin{tabular}{llr}
    \toprule
    \textbf{Term} & \textbf{Condition} & \textbf{Value} \\
    \midrule
    $r_{\text{format}}$ 
      & Valid JSON + thinking field   & $+0.5$ \\
      & Unparseable JSON              & $-3.0$ \\
    \midrule
    $r_{\text{tool}}$   
      & Correct tool for intent       & $+1.0$ \\
      & Wrong / hallucinated tool     & $-1.0$ / $-1.5$ \\
    \midrule
    $r_{\text{env}}$    
      & \texttt{lookup\_account} first & $+0.5$ \\
      & Correct domain tool (verified) & $+2.0$ \\
      & \texttt{speak} per turn        & $-0.05$ \\
      & Milestone bonus               & $\le+1.05$ \\
      & Late-episode penalty ($s>7$)  & $-0.12(s{-}7)^{1.4}$ \\
      & Terminal judge $J$            & $\in[-2,+2]$ \\
    \midrule
    $r_{\text{end}}$    
      & End-call after full sequence  & $+1.0$ \\
    \bottomrule
  \end{tabular}
\end{table}

\subsection{Adversarial Arena}
\label{sec:arena}

Each arena match pits the current DA checkpoint against the 
current CA checkpoint, the most recently trained version of 
each agent. Each match runs $N{=}18$ episodes per scenario 
(10 scenarios, 180 episodes per match). $W_{\text{DA}}$ is 
computed over CA-winnable scenarios; safety-critical 
fraud-escalation scenarios are tracked separately as a 
reference check. At the end of each match the stopping 
criterion (\S\ref{sec:stop}) is evaluated; if training 
continues, the match outcome determines which agent enters 
the Isolation Gym (\S\ref{sec:gym}), where it retrains 
against a frozen opponent before the next match begins.

\subsection{Isolation Gym}
\label{sec:gym}

If the stopping criterion has not fired, the weaker agent 
enters the Isolation Gym while its opponent's weights remain 
frozen. The gym constructs a training set from recent arena 
transcripts using a 70:30 failure/success mix, targeting 
failure modes from the most recent match while retaining 
enough successes to prevent catastrophic forgetting. A 
sliding-window slicer walks each transcript turn by turn: at 
every agent decision point, all prior dialogue accumulates as 
a prompt, yielding one GRPO training sample per turn depth.

The DA and CA gyms share this structure but differ in three 
ways reflecting their asymmetric action spaces: (i) $K{=}16$ 
generations per step for the DA versus $K{=}8$ for the CA, 
since structured JSON output requires greater rollout 
diversity; (ii) context and completion lengths are halved for 
the CA, whose utterances are 1--2 natural-language sentences 
rather than structured JSON; and (iii) reward functions differ entirely, with the CA gym
optimising $r_{\text{format}} + r_{\text{realism}} +
r_{\text{conceal}} + r_{\text{inv-DA}}$ rather than the DA's
tool-routing rewards. Full hyperparameters 
are in Appendix~\ref{app:training_config}. 
% The transcript mix 
% ratio, gym step budget, and rollout counts were chosen from 
% early stability observations rather than a systematic sweep; 
% sensitivity analysis is left as a dedicated future study.

\subsection{Stopping Criterion}
\label{sec:stop}
The stopping criterion is evaluated at the end of each arena 
match, using the same frozen opponent checkpoints that the 
% match was played against. Training terminates when both a 
% competency gate 
% ($W_{\text{DA}}^{(t)} \ge \tau_{\text{DA}}{=}0.60$) and a 
% stability gate 
% ($|W_{\text{DA}}^{(t)} - W_{\text{DA}}^{(t-1)}| \le 
% \varepsilon{=}0.05$) pass for $P{=}2$ consecutive matches 
% after $n_{\min}{=}3$, or when $n_{\max}{=}8$ matches are 
% exhausted
Training terminates when the competency gate 
($W_{\text{DA}}^{(t)} \ge \tau_{\text{DA}}{=}0.60$) holds for 
$P{=}2$ consecutive matches and the stability gate 
($|W_{\text{DA}}^{(t)} - W_{\text{DA}}^{(t-1)}| \le 
\varepsilon{=}0.05$) holds on the most recent of those matches, 
after $n_{\min}{=}3$, or when $n_{\max}{=}8$ matches are 
exhausted. The win-rate convergence criterion is an empirical 
surrogate for exploitability (NashConv)~\cite{lanctot2017unified}: 
$W_{\text{DA}}$ against a frozen opponent approximates a 
one-shot best response value, and we use its inter-match 
stability as a practical proxy where exact best-response 
computation is intractable.

%%%%%%%%%%%%%%%%%%%%%%%%%%%%%%%%%%%%%%%%%%%%%%%%%%%%%%%%%%%%%%%%%%%%%%
\section{Experiments}
\label{sec:exp}

\subsection{Setup}

All training uses 4-bit LoRA on Qwen2.5-3B (DA) and 
Qwen2.5-7B (CA and Judge); full hyperparameters and 
training rationale are in Appendix~\ref{app:training_config}. 
Arena matches used 18 episodes per scenario across 10 
scenarios, averaged over 2 runs; held-out evaluation uses 
12 fixed scenarios.

We use the following checkpoint notation throughout. 
$\text{DA}_0$ is the baseline DA produced by SFT on 
Banking77 demonstrations followed by GRPO warm-up with 
automated judge reward; it enters the first arena match 
without any isolation gym training. $\text{CA}_0$ is the 
baseline CA trained from reward pressure against the 
fixed $\text{DA}_0$ checkpoint, with no SFT initialisation. 
Subsequent checkpoints $\text{DA}_1$, $\text{DA}_2$ and 
$\text{CA}_1$, $\text{CA}_2$ are produced by successive 
isolation gym cycles: each subscript increment represents 
one gym cycle completed by that agent.

\section{Results}
\label{sec:results}

\subsection{Arena Progression}
\label{sec:exp:arena}

\begin{figure}[t]
  \centering
  \includegraphics[width=\linewidth]{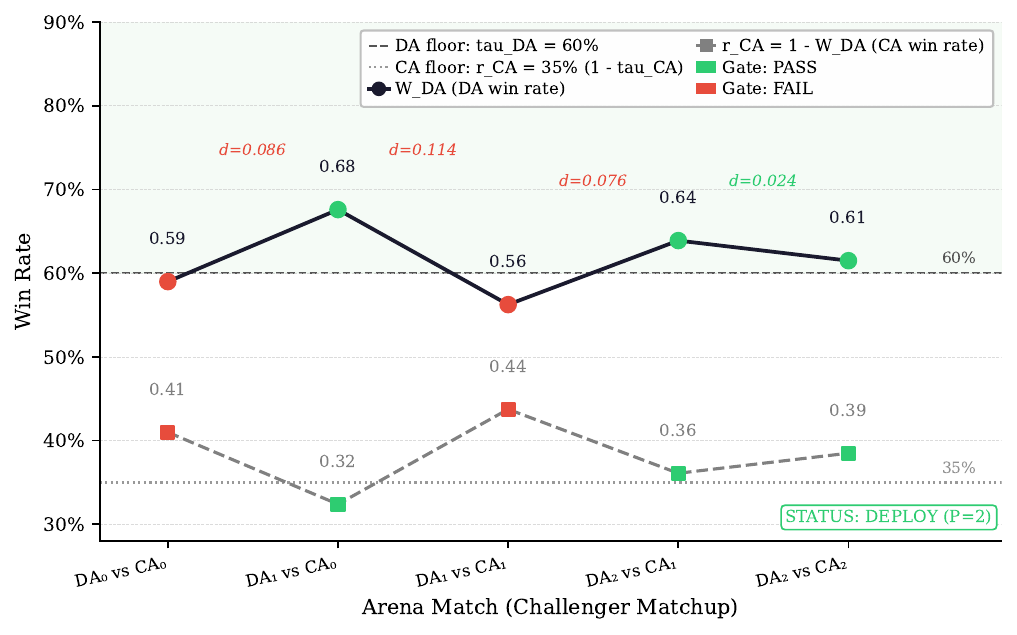}
  \caption{DA and CA win rates across five co-evolutionary arena 
           matches. $W_{\text{DA}}$ drops at matches 3 and 5 each 
           time the CA completes a gym cycle, consistent with 
           adversarial co-evolutionary pressure. Dashed reference 
           lines show competency floors 
           ($\tau_{\text{DA}}{=}0.60$, $r_{\text{CA}}{=}0.35$).}
  \label{fig:winrate}
\end{figure}

Table~\ref{tab:arena} and Figure~\ref{fig:winrate} show 
$W_{\text{DA}}$ across five matches. The trajectory is 
non-monotonic: $W_{\text{DA}}$ drops at matches 3 and 5, each 
time after the CA has completed a gym cycle, then recovers when 
the DA retrains. This pattern, visible in 
Figure~\ref{fig:winrate}, is consistent with genuine 
co-evolutionary pressure rather than independent improvement 
by each agent, though the small number of matches and high 
per-match variance ($\pm$8pp) limit how strongly this can be 
interpreted. The stopping criterion fires at match 5 
($|\delta|{=}0.02 \le \varepsilon$, $P{=}2$), with three of 
the eight maximum cycles unused. Per-scenario breakdown is in 
Appendix~\ref{app:breakdown}.

\begin{table}[t]
  \centering
  \caption{Arena match history. $W_{\text{DA}}$ is the mean over 2
           independent runs; differences $<{\sim}0.10$ are 
           directional only.}
  \label{tab:arena}
  \footnotesize
  \begin{tabular}{cllrrrc}
    \toprule
    \textbf{M} & \textbf{DA} & \textbf{CA}
      & $W_{\text{DA}}$ & $\sigma$
      & $\delta$ & \textbf{Gym} \\
    \midrule
    1 & $\text{DA}_0$ & $\text{CA}_0$ & 0.59 & 0.08 & ---     & --- \\
    2 & $\text{DA}_1$ & $\text{CA}_0$ & 0.68 & 0.08 & $+$0.09 & DA  \\
    3 & $\text{DA}_1$ & $\text{CA}_1$ & 0.56 & 0.08 & $-$0.12 & CA  \\
    4 & $\text{DA}_2$ & $\text{CA}_1$ & 0.64 & 0.08 & $+$0.08 & DA  \\
    5 & $\text{DA}_2$ & $\text{CA}_2$ & 0.62 & 0.08 & $-$0.02 & CA  \\
    \bottomrule
  \end{tabular}
\end{table}

\subsection{DA Improvement on Held-out Evaluation}

Table~\ref{tab:eval} compares $\text{DA}_0$ and $\text{DA}_2$ 
on the fixed held-out set ($n{=}12$; treat as indicative). 
PASS requires correct tool routing, \texttt{lookup\_account} 
called first, $r_{\text{DA}} \ge 2.0$, and a clean ending.

After five co-evolutionary cycles, all routing errors are 
eliminated: $\text{DA}_2$ selects the correct tool on every 
held-out scenario, up from 75\% in $\text{DA}_0$, and the strict 
PASS rate rises from 25\% to 50\%. These gains occur without 
labelled data; the sole training signal is automated arena reward. 
Although the aggregate FAIL rate is unchanged at 33\%, the failure 
mode shifts entirely: $\text{DA}_0$ fails through routing errors 
(calling \texttt{transfer\_to\_human} on dispute and transfer 
scenarios), while $\text{DA}_2$ routes correctly on all scenarios 
but fails on procedural and conversation-quality criteria. Routing failures are resolved; remaining procedural failures are addressable with additional gym cycles targeting 
conversation-quality criteria.

\begin{table}[t]
  \centering
  \caption{Held-out evaluation: $\text{DA}_0$ (baseline) vs.\
           $\text{DA}_2$ (final, after 5 co-evolutionary cycles).
           12 fixed scenarios, averaged over 2 runs.}
  \label{tab:eval}
  \footnotesize
  \begin{tabular}{lccc}
    \toprule
    \textbf{Metric} & $\mathbf{DA_0}$ & $\mathbf{DA_2}$ & $\Delta$ \\
    \midrule
    Tool-routing accuracy & 75\%  & 100\% & \textbf{+25pp} \\
    PASS rate (strict)    & 25\%  & 50\%  & \textbf{+25pp} \\
    FAIL rate             & 33\%  & 33\%  & 0              \\
    Avg episode reward    & $+1.58$ & $+2.16$ & \textbf{+0.58} \\
    Lookup-first rate     & 58\%  & 83\%  & \textbf{+25pp} \\
    \bottomrule
  \end{tabular}
\end{table}

\subsection{Emergent Behaviour: Contextual Camouflage}
\label{sec:exp:ca}

We classify the CA's opening utterance via manual inspection 
of sampled transcripts across $n{\approx}50$ episodes per 
match (Table~\ref{tab:tactics}). The primary emergent 
behaviour we observe is what we term \textbf{Contextual 
Camouflage}: trained CA models learn to embed the true intent 
within dense, realistic customer detail, citing specific 
merchant names, transaction amounts, and incident context.

\begin{table}[t]
  \centering
  \caption{CA opening-utterance tactic distribution, manual 
           classification, $n{\approx}50$ per match. 
           M4 omitted (same CA checkpoint as M3).}
  \label{tab:tactics}
  \footnotesize
  \begin{tabular}{lcccc}
    \toprule
    \textbf{Tactic} & \textbf{M1} & \textbf{M2} & 
                      \textbf{M3} & \textbf{M5} \\
    \midrule
    Direct intent    & 25 & 34 & 37 & 34 \\
    Vague/indirect   & 25 & 16 & 13 & 16 \\
    Emotional        &  2 &  4 &  2 &  2 \\
    Role reversal    &  0 &  0 &  0 &  0 \\
    \bottomrule
  \end{tabular}
\end{table}

This is \emph{distinct from vagueness}: Direct Intent 
increases (25$\to$34$\to$37) while Vague/Indirect decreases 
(25$\to$16$\to$13). The CA is not becoming more vague; it is 
becoming more \emph{specifically misleading}, naming the 
correct domain but layering it in contextual noise that forces 
the DA to arbitrate between competing signals simultaneously. 
This behaviour emerges purely from the reward signal of 
maximising DA misroutes, without any explicit specification. 
We treat this as a qualitative observation; future work will 
measure utterance specificity directly via named-entity density 
or opening-turn token length.

%%%%%%%%%%%%%%%%%%%%%%%%%%%%%%%%%%%%%%%%%%%%%%%%%%%%%%%%%%%%%%%%%%%%%%
\section{Limitations}
\label{sec:limits}

\begin{itemize}
\setlength{\itemsep}{1pt}
  \setlength{\parsep}{0pt}
  \setlength{\topsep}{2pt}

  \item \textbf{Stopping criterion generality.} The macro 
        stopping criterion uses two gates (competency and 
        stability) with fixed thresholds chosen pragmatically; 
        production deployment requires richer criteria combining 
        non-negotiable scenario coverage, domain-specific 
        compliance gates, and customer-segment performance 
        floors. Generalising the stopping criterion to a 
        configurable per-domain gating framework is left to 
        future work.

  \item \textbf{Evaluation scope.} All experiments are in a 
        single banking domain with a small number of matches 
        and held-out scenarios; cross-domain generalisation 
        and finer-grained statistical separation are not 
        established.

  \item \textbf{Ablations and sensitivity.} Relative contributions 
      of reward components are unquantified; reward weights, 
      stopping thresholds $\tau_{\text{DA}}$, $\varepsilon$, 
      $P$, the 70:30 transcript mix, and gym hyperparameters 
      were chosen pragmatically; systematic sweeps and reward 
      ablations are left to future work.

  \item \textbf{Live-deployment transfer.} The data flywheel is 
      demonstrated within arena training, not against live 
      deployment outcomes. Closing the loop in production 
      requires a deployment time outcome signal that 
      substitutes for the judge's terminal score; in practice 
      this maps onto existing enterprise telemetry 
      (resolution markers, callback rates, escalation traces) 
      rather than new annotation.

  \item \textbf{Role-reversal under prompt mitigation.} CA 
        outputs occasionally drift toward bank-agent 
        impersonation; this is mitigated through stronger 
        system prompts (Appendix~\ref{app:prompts}) but not 
        provably under longer training. A role-consistency 
        reward term is the principled fix and is left to 
        future work.
\end{itemize}

\section{Production Deployment Considerations}
\label{sec:production}

\paragraph{Toward a post-deployment data flywheel.}
The Isolation Gym provides the training-side mechanism for 
an enterprise data flywheel: failure transcripts feed 
retraining directly via a 70:30 failure/success mix, with no 
human annotation. Closing the loop end-to-end in production 
requires a deployment-time outcome signal (resolution markers, 
callback rates, agent escalations) that substitutes for the 
judge's terminal score; this is left to future work and 
discussed in Limitations.

\paragraph{Deployment gating.}
The macro stopping criterion in \S\ref{sec:stop} signals 
that co-evolution has stabilised. Production deployment 
requires a stronger gate: non-negotiable scenarios (fraud 
detection, account compromise, identity verification) must 
pass a higher threshold (e.g.\ $\ge 90\%$ DA win rate 
against the latest CA) before promotion. This deployment 
gate is orthogonal to the training stopping criterion and 
allows enterprise operators to enforce compliance-critical 
behaviours independently of overall convergence.

\paragraph{Domain transferability.}
RL-ADA's components are domain-agnostic: the adversarial 
arena, isolation gym, and world-feedback reward require only 
that interaction outcomes are measurable. The 78-intent 
banking taxonomy can be replaced with any domain-specific 
intent-to-tool map, and the judge can be re-validated against 
a domain-appropriate reference. The deployment-gating 
threshold and non-negotiable scenario set are configured 
per-domain to reflect each enterprise's compliance and 
escalation cost structure.

%%%%%%%%%%%%%%%%%%%%%%%%%%%%%%%%%%%%%%%%%%%%%%%%%%%%%%%%%%%%%%%%%%%%%%
\section{Conclusion}
\label{sec:conclusion}

RL-ADA trains a 3B Customer Support Agent against a 7B 
Adversarial Customer Agent using only world feedback (measurable
conversation outcomes) with no human annotation. Over five
co-evolutionary cycles, all routing errors are eliminated on a
12-scenario held-out benchmark, the strict PASS rate doubles 
from 25\% to 50\%, and average episode reward rises from 
$+1.58$ to $+2.16$, driven solely by automated arena reward 
with no labelled data. A macro-level stopping criterion based 
on win-rate stability identifies convergence at cycle 5, with 
three of eight maximum cycles unused. Two findings stand out 
beyond the headline metrics. First, the arena win-rate 
trajectory is non-monotonic: DA performance drops whenever 
the CA completes an isolation gym cycle, consistent with 
genuine adversarial co-evolutionary pressure. Second, the 
trained CA develops \emph{Contextual Camouflage}, embedding 
intent within dense, specific customer detail rather than 
naming it directly; this adversarial behaviour emerged from 
reward pressure alone, without explicit specification.

These are preliminary results on a single banking domain. 
RL-ADA's components are designed for cross-domain transfer: 
world-feedback rewards, opposing-reward co-evolution, and the 
isolation gym training mechanism each depend only on 
measurable interaction outcomes rather than domain-specific 
labels. Demonstrating this transfer empirically is the 
immediate next step.

%%%%%%%%%%%%%%%%%%%%%%%%%%%%%%%%%%%%%%%%%%%%%%%%%%%%%%%%%%%%%%%%%%%%%%
%% bibliography
\bibliographystyle{ACM-Reference-Format}
\bibliography{rlada_references}

%%%%%%%%%%%%%%%%%%%%%%%%%%%%%%%%%%%%%%%%%%%%%%%%%%%%%%%%%%%%%%%%%%%%%%
\appendix

\section{Intent-to-Tool Taxonomy}
\label{app:taxonomy}

Table~\ref{tab:taxonomy} shows how Banking77 intents are grouped
into the six routing tools used in RL-ADA. The mapping is
many-to-one: 78 intent strings (Banking77 plus 6 core synthetic
labels) are collapsed to 6 tools. The large
\texttt{transfer\_to\_human} bucket (38 intents) reflects that
Banking77 contains many account-management and policy intents that
have no automated resolution path. Intents in the core set
($\dagger$) are used in all arena scenarios; Banking77 extensions
provide vocabulary diversity during SFT and GRPO warm-up.

\begin{table}[h]
  \centering
  \caption{Intent-to-tool routing map. $n$: number of intent strings
           per tool. $\dagger$: core intent.}
  \label{tab:taxonomy}
  \footnotesize
  \begin{tabular}{llp{3cm}}
    \toprule
    \textbf{Tool} & $n$ & \textbf{Example intents} \\
    \midrule
    \texttt{lookup\_account}     &  2 &
      \texttt{check\_balance}$^\dagger$,
      \texttt{verify\_my\_identity} \\
    \texttt{get\_transactions}   &  8 &
      \texttt{check\_transactions}$^\dagger$ \\
    \texttt{dispute\_charge}     & 19 &
      \texttt{dispute\_charge}$^\dagger$,
      \texttt{unrecognized\_charge}$^\dagger$ \\
    \texttt{check\_card\_status} &  9 &
      \texttt{card\_declined}$^\dagger$,
      \texttt{pin\_blocked} \\
    \texttt{transfer\_funds}     &  2 &
      \texttt{transfer\_funds}$^\dagger$ \\
    \texttt{transfer\_to\_human} & 38 &
      \texttt{fraud\_alert}$^\dagger$,
      \texttt{compromised\_card} (+35) \\
    \midrule
    \textbf{Total}               & \textbf{78} & 6 routing targets \\
    \bottomrule
  \end{tabular}
\end{table}

The skew toward \texttt{transfer\_to\_human} creates a class
imbalance: a DA that over-generalises to escalation scores well on
coverage but fails on the five more specific tools. This is the
failure mode $\text{DA}_0$ exhibits, and the arena's CA-winnable
scenario set is designed to stress-test the non-escalation buckets.

\section{Training Configurations}
\label{app:training_config}

\subsection{DA Training: Baseline vs.\ Isolation Gym}

$\text{DA}_0$ was trained before the isolation gym protocol was
standardised and used a higher learning rate and more steps.
$\text{DA}_1$ and $\text{DA}_2$ were produced by the standardised
isolation gym with the parameters in Table~\ref{tab:config_da}.

\begin{table}[h]
  \centering
  \caption{DA training configurations.}
  \label{tab:config_da}
  \footnotesize
  \begin{tabular}{lll}
    \toprule
    \textbf{Parameter} & \textbf{DA$_0$ (baseline)} &
    \textbf{DA$_{1,2}$ (gym)} \\
    \midrule
    Base model           & Qwen2.5-3B-Instruct & Qwen2.5-3B-Instruct \\
    LoRA rank / $\alpha$ & 16 / 16             & 16 / 16 \\
    Learning rate        & $2{\times}10^{-5}$  & $5{\times}10^{-6}$ \\
    LR schedule          & const.\ w/ warmup   & const.\ w/ warmup \\
    KL penalty $\beta$   & 0.001               & 0.02 \\
    GRPO generations $K$ & 16                  & 16 \\
    Batch size           & 1 (grad accum 4)    & 1 (grad accum 4) \\
    Training steps       & 300                 & 80 \\
    \bottomrule
  \end{tabular}
\end{table}

\subsection{CA Isolation Gym}

All CA gym cycles (producing $\text{CA}_1$ and $\text{CA}_2$) 
used the parameters in Table~\ref{tab:config_ca}. The CA uses 
cosine LR and fewer generations per step than the DA gym, 
reflecting the shorter utterance space and a tighter KL 
constraint ($\beta{=}0.05$) to keep outputs within 
natural-language register.

\begin{table}[h]
  \centering
  \caption{CA isolation gym configuration (all cycles).}
  \label{tab:config_ca}
  \footnotesize
  \begin{tabular}{ll}
    \toprule
    \textbf{Parameter} & \textbf{Value} \\
    \midrule
    Base model           & Qwen2.5-7B-Instruct \\
    LoRA rank / $\alpha$ & 16 / 16 \\
    Learning rate        & $5{\times}10^{-6}$ \\
    LR schedule          & cosine \\
    KL penalty $\beta$   & 0.05 \\
    GRPO generations $K$ & 8 \\
    Batch size           & 1 (grad accum 4) \\
    Training steps       & 80 \\
    Dataset mix          & 70:30 failure/success \\
    \bottomrule
  \end{tabular}
\end{table}

\section{Sample Episode Transcripts}
\label{app:transcripts}

\subsection{Contextual Camouflage --- CA Wins by Burying Intent}

\noindent{\footnotesize\textbf{Scenario:} dispute-duplicate-v2
\textbf{|} Expected tool: \texttt{dispute\_charge}}

\begin{quote}\small
\textbf{CA ($\text{CA}_2$):} \textit{``It seems like I might have
accidentally ordered two drinks or there could be some confusion
with my payment.''}

\noindent\textbf{DA ($\text{DA}_2$):}
[\texttt{lookup\_account}] $\to$ [\texttt{transfer\_to\_human}]
\textit{(wrong)}
\end{quote}

\noindent The CA embeds the dispute intent behind self-doubt phrasing;
the DA escalates instead of filing a dispute.

\subsection{DA Improvement --- Before and After Training}

\noindent{\footnotesize\textbf{Utterance:} \textit{``I see a charge
I don't recognise from `AMZN MKTP' for \$89.''}}

\begin{quote}\small
\textbf{$\text{DA}_0$:} [\texttt{transfer\_to\_human}]
\hfill\textit{(wrong)}

\smallskip
\textbf{$\text{DA}_2$:} [\texttt{lookup\_account}] $\to$
[\texttt{dispute\_charge}] $\to$ \textit{``I have disputed the
charge\ldots''} $\to$ \texttt{end\_call}
\hfill\textit{(correct)}
\end{quote}

\noindent $\text{DA}_2$ correctly identifies the dispute intent and
closes the call without any new labelled examples.

\section{System Prompts}
\label{app:prompts}

\subsection{DA System Prompt (SFT + GRPO)}

\begin{quote}\small\ttfamily
You are a professional banking customer support agent.

AVAILABLE TOOLS:\\
\ \ lookup\_account\ \ \ \ {---} verify customer identity\\
\ \ get\_transactions\ \ {---} retrieve last 5 transactions\\
\ \ dispute\_charge\ \ \ \ {---} file a dispute\\
\ \ check\_card\_status\ {---} check/unblock a declined card\\
\ \ transfer\_funds\ \ \ \ {---} move money between accounts\\
\ \ transfer\_to\_human\ {---} escalate to specialist

CALL SEQUENCE (each step ONCE):\\
\ \ 1.\ lookup\_account\ \ (always first)\\
\ \ 2.\ [domain tool]\ \ \ (one, based on customer need)\\
\ \ 3.\ speak\ \ \ \ \ \ \ \ \ \ \ (explain result in plain English)\\
\ \ 4.\ end\_call\ \ \ \ \ \ \ \ (when customer is satisfied)

FORMAT (always valid JSON):\\
\{"thinking": "[LAST ACTION:...][HAVE:...]->Next:...",\\
\ "action\_type": "speak|tool\_call|end\_call",\\
\ "content": "message OR tool name", "tool\_args": \{\}\}
\end{quote}

\subsection{CA System Prompt}

\begin{quote}\small\ttfamily
You are a banking customer calling support.
You are NOT the agent.

- NEVER offer to help or diagnose.\\
- Always use "I" as someone confused or frustrated.\\
- You called THEM. Wait for THEM to help YOU.\\
- Express your problem naturally and indirectly.\\
- Keep replies to 1--2 sentences.
\end{quote}

\section{Per-Scenario Match 5 Breakdown}
\label{app:breakdown}

\begin{figure}[h]
  \centering
  \includegraphics[width=\linewidth]{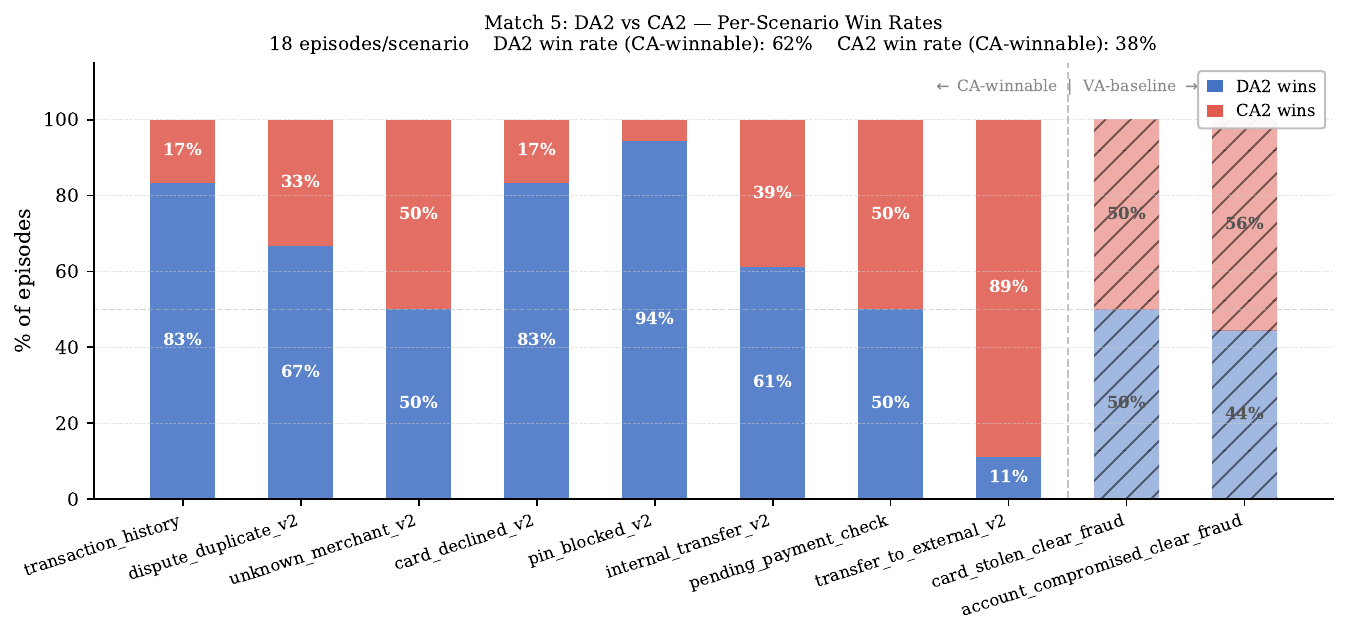}
  \caption{Per-scenario win rates for Match 5, 18 episodes per 
           scenario. Solid bars are CA-winnable scenarios; 
           hatched bars are DA-baseline fraud scenarios.}
  \label{fig:breakdown}
\end{figure}

Figure~\ref{fig:breakdown} shows per-scenario win rates for 
Match 5 ($\text{DA}_2$ vs $\text{CA}_2$). CA pressure 
concentrates on ambiguous-intent scenarios 
(\texttt{dispute\_duplicate}, \texttt{unknown\_merchant}, 
\texttt{internal\_transfer}); $\text{DA}_2$ retains near-perfect 
accuracy on clear-intent scenarios.

\end{document}